\documentclass{article}
\usepackage[final]{neurips_mlsb_2025}
\usepackage{graphicx}
\usepackage{amsmath, amssymb}

\usepackage{booktabs}  
\usepackage{url}
\usepackage{tabularx}
\usepackage{float}
\usepackage{svg}
\usepackage{xcolor}
\usepackage{hanging}
\usepackage{ragged2e}   
\usepackage{subcaption}
\usepackage[font=small,skip=0.3em]{caption}
\usepackage{subcaption}

\usepackage[utf8]{inputenc} 
\usepackage[T1]{fontenc}    
\usepackage{hyperref}       
\usepackage{url}            
\usepackage{booktabs}       
\usepackage{amsfonts}       
\usepackage{nicefrac}       
\usepackage{microtype}      
\usepackage{xcolor}         
\usepackage{wrapfig}
\usepackage{adjustbox}

\usepackage{titlesec}

\titlespacing*{\section}
  {0pt}{1.0ex plus 0.2ex minus 0.2ex}{0.8ex plus 0.1ex}
\titlespacing*{\subsection}
  {0pt}{0.8ex plus 0.2ex minus 0.2ex}{0.5ex plus 0.1ex}
\titlespacing*{\subsubsection}
  {0pt}{0.5ex plus 0.1ex minus 0.1ex}{0.3ex plus 0.1ex}

\title{\textit{De novo} generation of functional terpene synthases using TpsGPT}

\author{
    Hamsini Ramanathan\textsuperscript{1} \quad
    Roman Bushuiev\textsuperscript{2, 3} \quad
    Matouš Soldát\textsuperscript{3} \quad
    Jiří Kohout\textsuperscript{3} \\
    \textbf{Téo Hebra}\textsuperscript{3} \quad
    \textbf{Joshua David Smith}\textsuperscript{3} \quad
    \textbf{Josef Sivic}\textsuperscript{2} \quad
    \textbf{Tomáš Pluskal}\textsuperscript{3} \\
    \\
    \textsuperscript{1}Seattle Academy of Arts and Sciences (SAAS), Seattle \\[1ex]
    \textsuperscript{2}Czech Institute of Informatics, Robotics and Cybernetics (CIIRC), Czech Technical University\\[1ex]
    \textsuperscript{3}Institute of Organic Chemistry and Biochemistry of the Czech Academy of Sciences \\[1ex]
}
\date{August 2025}

\begin{document}

\maketitle

\begin{abstract}

Terpene synthases (TPS) are a key family of enzymes responsible for generating the diverse terpene scaffolds that underpin many natural products, including front-line anticancer drugs such as Taxol. However, \textit{de~novo} TPS design through directed evolution is costly and slow. We introduce TpsGPT, a generative model for scalable TPS protein design, built by fine-tuning the protein language model ProtGPT2 on 79k TPS sequences mined from UniProt. TpsGPT generated \textit{de novo} enzyme candidates \textit{in silico} and we evaluated them using multiple validation metrics, including EnzymeExplorer classification, ESMFold structural confidence~(pLDDT), sequence diversity, CLEAN classification, InterPro domain detection, and Foldseek structure alignment. From an initial pool of 28k generated sequences, we identified seven putative TPS enzymes that satisfied all validation criteria. Experimental validation confirmed TPS enzymatic activity in at least two of these sequences. Our results show that fine-tuning of a protein language model on a carefully curated, enzyme-class-specific dataset, combined with rigorous filtering, can enable the \textit{de~novo} generation of functional, evolutionarily distant enzymes.

\end{abstract}

\section{Introduction}
Terpene synthases~(TPS) are a specialized family of enzymes that generate hydrocarbon scaffolds for terpenes—the largest and most diverse class of natural products, encompassing widely used flavors, fragrances, and frontline medicines~\citep{samusevich-2024}. Terpenes exhibit diverse bioactivities, including analgesic, anticonvulsant, and anti-inflammatory properties~\citep{del-prado-audelo-2021}. More than 76,000 terpenes have been characterized to date~\citep{rudolf-2019}. Among them, Taxol, a diterpene, remains a first-line anticancer drug with multi-billion-dollar annual sales~\citep{weaver-2014}. 

Despite their importance, terpenes are notoriously difficult to synthesize industrially due to their structural complexity~\citep{del-moral-2019}. Conventional chemical synthesis requires numerous steps and incurs high energy and resource costs, making it unsustainable at scale. In contrast, synthetic biology offers a more efficient route by leveraging TPS enzymes to catalyze key reactions~\citep{zhang-2020}.

Here we present \textbf{TpsGPT}\footnote{\url{https://github.com/colorfulcereal/TpsGPT}}, a terpene synthase sequence generation model fine-tuned on a distilled protein language model --- ProtGPT2 Tiny~\citep{ferruz-2022, protgpt2-tiny}. TpsGPT is trained on a carefully curated 79k homologous TPS dataset mined from large scale repositories like UniProt. The mining process initially used a very small \textbf{1125} experimentally characterized actual TPS enzymes from published sources as a seed to identify TPS patterns based on which the mining process produced 79k homologous TPS sequences. TpsGPT generated evolutionary distant sequences while conserving key TPS structural features. The resulting \textit{de novo} sequences exhibit high predicted structural stability and low sequence identity relative to the training set. Our results demonstrate that fine-tuning protein language models on a carefully curated, enzyme-class-specific dataset, can effectively explore the vast protein sequence space, producing valid enzyme candidates even for underrepresented protein families like terpene synthases.

\section{Related Work}


\paragraph{Protein engineering.} The design of novel TPS enzymes for terpene biosynthesis remains a complex and time-consuming task. There are two main approaches for protein engineering: rational design and directed evolution~\citep{vidal-2023}. Rational design involves performing chosen point mutations, insertions or deletions in the coding sequence. Directed evolution, on the other hand, bypasses the need to determine specific mutations a priority by mimicking the process of natural evolution in the laboratory. While promising, these methods have a major disadvantage --- the sequences they generate often remain highly similar to naturally occurring proteins, leaving vast regions of the protein sequence space unexplored~\citep{yang-2024B}. Moreover, robotics-accelerated high-throughput directed evolution techniques like Phage-Assisted Continuous Evolution are prohibitively expensive, with costs reaching hundreds of thousands of dollars~\citep{aoudjane-2024}

\paragraph{Computational design of terpene synthases.} Machine learning-assisted annotation methods predict and label likely TPS enzymes in large protein databases like UniProt and UniRef~\citep{samusevich-2024, alex-2018, suzek-2014} but such methods only uncover existing proteins in nature. \textit{De novo} enzyme design approaches such as RFdiffusion use diffusion-based deep learning architectures to generate novel protein backbones~\citep{watson-2023}. Although promising, RFDiffusion is a structure-based method and requires a comprehensive understanding of a catalytic site and its activity to generate functional enzymes~\citep{lauko2025computational}. Instead, our work aims to design enzymes given only a set of family-specific sequences.


\paragraph{Protein Language Models (PLMs).} PLMs are based on large language models (LLMs) like GPT2, which leverage the Transformer architecture to model sequential data~\citep{vaswani-2017}. Prior work has shown that fine-tuning PLMs can generate \textit{de novo} proteins within specific families~\citep{winnifrith-2024}. However, existing PLM fine-tuning methods to generate sequences rely on extensive family-specific datasets and often require additional inputs such as control tags for model conditioning. Additionally, the fine-tuning is typically done on large models such as ProGEN with 280 million parameters~\citep{madani-2023}. ProtGPT2 is a state-of-the-art autoregressive Transformer-based PLM with 738 million parameters~\citep{ferruz-2022} and enables high-throughput protein generation in seconds. Additionally, ProtGPT2 offers a tiny model~\citep{protgpt2-tiny} with 38.9 million parameters with comparable performance as the original bigger model. Motivated by these properties, we fine-tune ProtGPT2 tiny to generate \textit{de novo} terpene synthase sequences starting from a small dataset of 1125 TPS sequences.

\vspace{-0.25cm}

\section{Materials and Methods}

\begin{figure}[!t]
  \centering
  \includegraphics[width=1\textwidth]{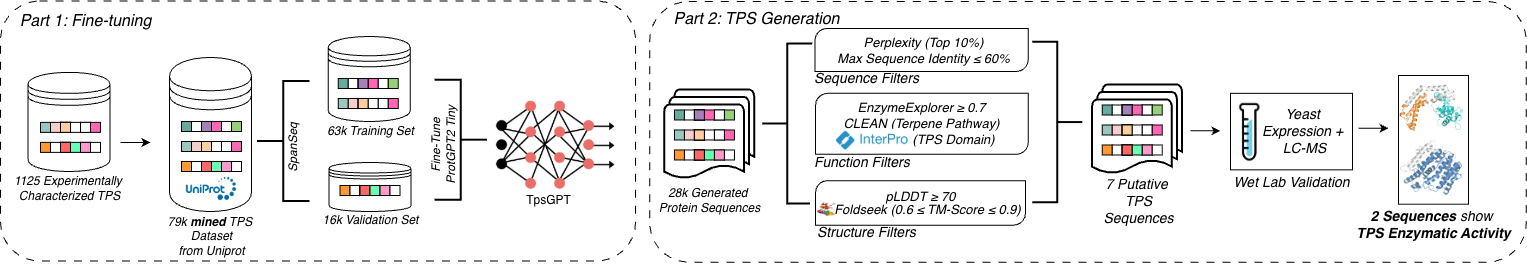}
  \caption{\textbf{Overview of our approach.}\textbf{ Part 1:} We collected 1125 experimentally characterized TPS enzymes from all published sources to mine the 79k TPS dataset from UniProt~\citep{pluskal-2024}. We created an 80/20 split using SpanSeq~\citep{florensa-2024} with at most 30\% sequence identity between the splits and fine-tuned the distilled ProtGPT2 tiny~\citep{protgpt2-tiny} model to create TpsGPT. \textbf{Part 2:} We generated 28k sequences using TpsGPT and filtered them using seven validation metrics: \textbf{Sequence filters:} Perplexity and max sequence identity to training set. \textbf{Function filters:} EnzymeExplorer TPS score~\citep{samusevich-2024}, CLEAN enzyme classification~\citep{yu-2023}, and InterPro domain prediction~\citep{blum-2024} \textbf{Structure filters:} pLDDT using ESMFold~\citep{esmfold} and max Foldseek TM-score to training set~\citep{van-kempen-2023}. Above filters reduced the 28k sequences to \textbf{seven} putative TPS sequences. Wet-lab validation using yeast expression followed by liquid chromatography coupled with mass spectrometry (LC-MS) showed TPS enzymatic activity in two sequences~\citep{pitt-2009}.}
  \label{fig:materials_methods}
\end{figure}

We developed \textbf{TpsGPT}, a scalable \textit{in silico} framework for \textit{de novo} TPS enzyme design (Figure \ref{fig:materials_methods}). The approach combines protein language model fine-tuning with principled sequence generation and multi-stage validation to produce viable, evolutionarily distant TPS candidates.

\subsection{Dataset Preparation}


As a starting point, we used a dataset of 1125 experimentally validated TPS sequences, which was later extended to 79k computationally-mined TPS sequences~\citep{pluskal-2024}. In the mining process, HMMER’s hmmsearch was used with merged and indexed TPS-specific Pfam and SUPERFAMILY databases to identify terpene synthases~\citep{finn-2011,mistry-2020,wilson-2006}. For the preclustered Big Fantastic Database (BFD), only representative sequences were mined first, with full clusters analyzed if hits were found~\citep{jumper-2021}. Post-mining, several filtering steps enhanced reliability: removing sequences outside the 300–1100 amino acid range, excluding those with stronger hits to non-TPS domains, requiring characteristic TPS catalytic motifs (DDXXD and NSE/DTE motifs of Class I TPSs and the DXDD motif of Class II TPSs), discarding sequences with incomplete or atypical domain architectures, and filtering out sequences with >80\% identity to known isoprenyl diphosphate synthases~\citep{liu-2025, jiang-2019}. The remaining sequences were designated as candidate terpene synthases and were used to fine-tune ProtGPT2.

To avoid data leakage and ensure generalization, the mined TPS sequences were clustered using SpanSeq~\citep{florensa-2024} into six partitions at 30\% sequence identity between partitions. We combined five partitions ($\sim$63k sequences) for training while the remaining partition ($\sim$16k sequences) was reserved for validation, resulting in an 80/20 split.

\vspace{-0.25em}

\subsection{Model Fine-Tuning}
The original ProtGPT2 model contains 738 million parameters, making full fine-tuning computationally expensive~\citep{nferruz-protgpt2}. Hence, we fine-tuned the distilled ProtGPT2 tiny model with 38.9 million parameters. The distilled tiny model retains comparable perplexities to the original large model while offering up to six times faster inference, enabling high-throughput sequence generation~\citep{protgpt2-tiny}. Fine-tuning was performed using Lightning AI on a single NVIDIA L4 tensor core~\citep{lightning-ai}.

\subsection{TPS Sequence Generation and Filtering}

After fine-tuning, we generated 28k protein sequences. A multi-stage filtering pipeline was applied to identify putative TPS enzymes from the 28k sequences:

    \textbf{Sequence Filters:} The 28k sequences were ranked by perplexity, and the top 10\% (2,800 sequences) were retained. Maximum pairwise sequence identity (maxID) to the training set was computed, and only sequences with maxID $\leq 60\%$ were retained to encourage evolutionary distance.
    
    \textbf{Function Filters:} We used EnzymeExplorer with a TPS detection threshold of 0.7 (range 0–1) to select sequences likely to possess TPS activity~\citep{samusevich-2024}. CLEAN (Contrastive Learning Enabled Enzyme ANnotation) is a ML model that assigns EC (Enzyme Commission) number to protein sequences~\citep{yu-2023}. We used CLEAN to predict EC numbers and selected only those with a terpenoid/terpene biosynthetic pathway in BRENDA~\citep{brenda-2025}. InterPro is another model that predicts domains given a sequence~\citep{blum-2024}. We selected only those sequences when the InterPro predicted domain was a terpene synthase specific domain or a domain with an overlapping superfamily containing a TPS domain.
    
    \textbf{Structure Filters:} We computed Predicted Local Distance Difference Test (pLDDT) scores from ESMFold~\citep{esmfold} and retained only sequences with pLDDT $\geq 70$ indicative of accurate backbone modeling and valid 3D structures. To ensure conservation of TPS structure, we used Foldseek to do structural comparison of the generated sequences relative to their respective top structural matches in the training set and retained those with TM-scores between 0.6 and 0.9~\citep{van-kempen-2023}. 
    
    This pipeline produced candidates that are structurally feasible, TPS-like, and evolutionarily distant, representing potential \textit{de novo} TPS enzymes suitable for downstream experimental validation.


\section{Results}

We fine-tuned the distilled ProtGPT2 tiny model on 79k TPS sequences mined from UniProt and generated 28k TPS sequence candidates. After picking the top 10\% (2800) sequences by perplexity score, we applied the following filters: pLDDT score, EnzymeExplorer TPS detection score, max sequence identity to training set, Foldseek alignment (TM-Score), CLEAN classified EC number, and InterPro domain to identify putative \textit{de novo} TPS sequences.

\subsection{TpsGPT Generates Valid TPS Sequences}

\begin{wrapfigure}{r}{0.5\textwidth}
\vspace{-1cm}
    \centering
    \includegraphics[width=\linewidth]{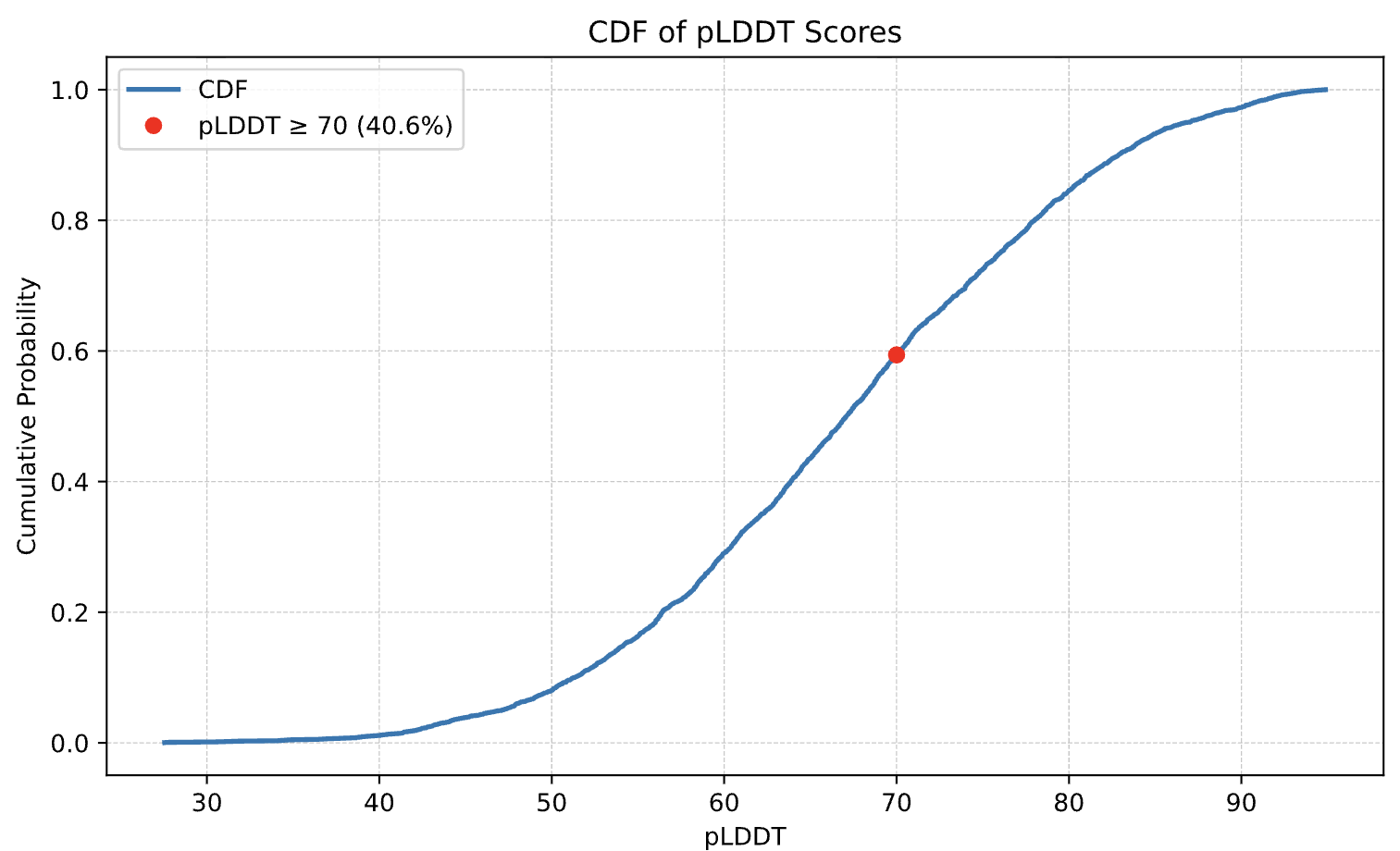}
    \caption{CDF of pLDDT scores of the top 2800 generated sequences. 
     More than 40\% had pLDDT $\geq$ 70 indicating stable structures.}
    \label{fig:pLDDT_scores}    
\vspace{-1cm}
\end{wrapfigure}

Among the top 2,800 sequences ranked by perplexity, 40\% achieved pLDDT scores $\geq 70$ (Figure~\ref{fig:pLDDT_scores}). From this set, \textbf{77 sequences} passed the EnzymeExplorer TPS detection threshold ($>=0.7$). A detection score above 0.7 indicates the potential to catalyze terpenes. 

\subsection{Evolutionarily Distant Sequences with Conserved TPS Structures}
\begin{figure}[H]
  \centering
  \begin{center}
    \begin{subfigure}[b]{0.16\textwidth}
        \includegraphics[width=\linewidth]{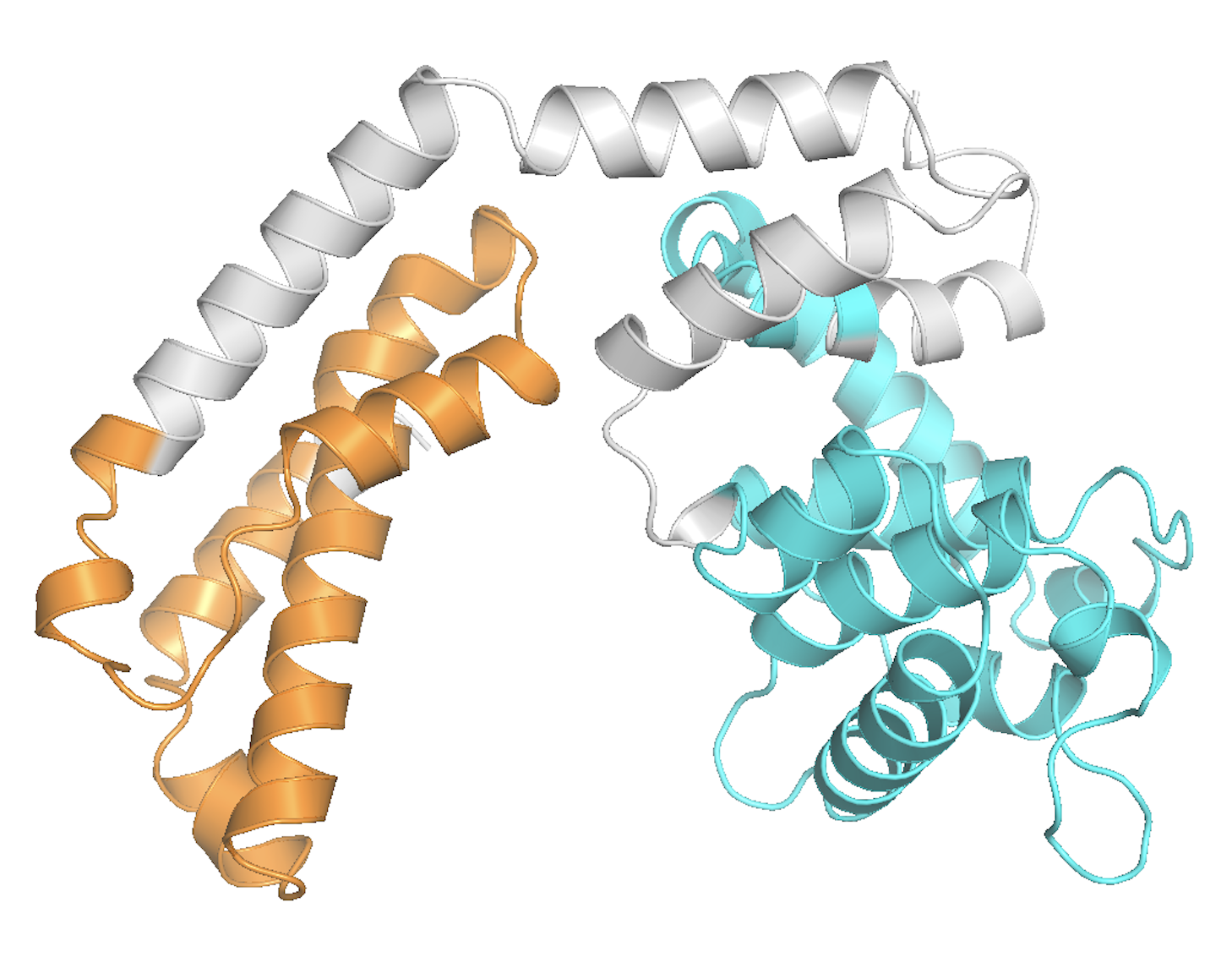}
        \caption*{TpsGPT1}
        \caption*{maxID 49.67\%}
    \end{subfigure}\hfill
    \begin{subfigure}[b]{0.16\textwidth}
        \includegraphics[width=\linewidth]{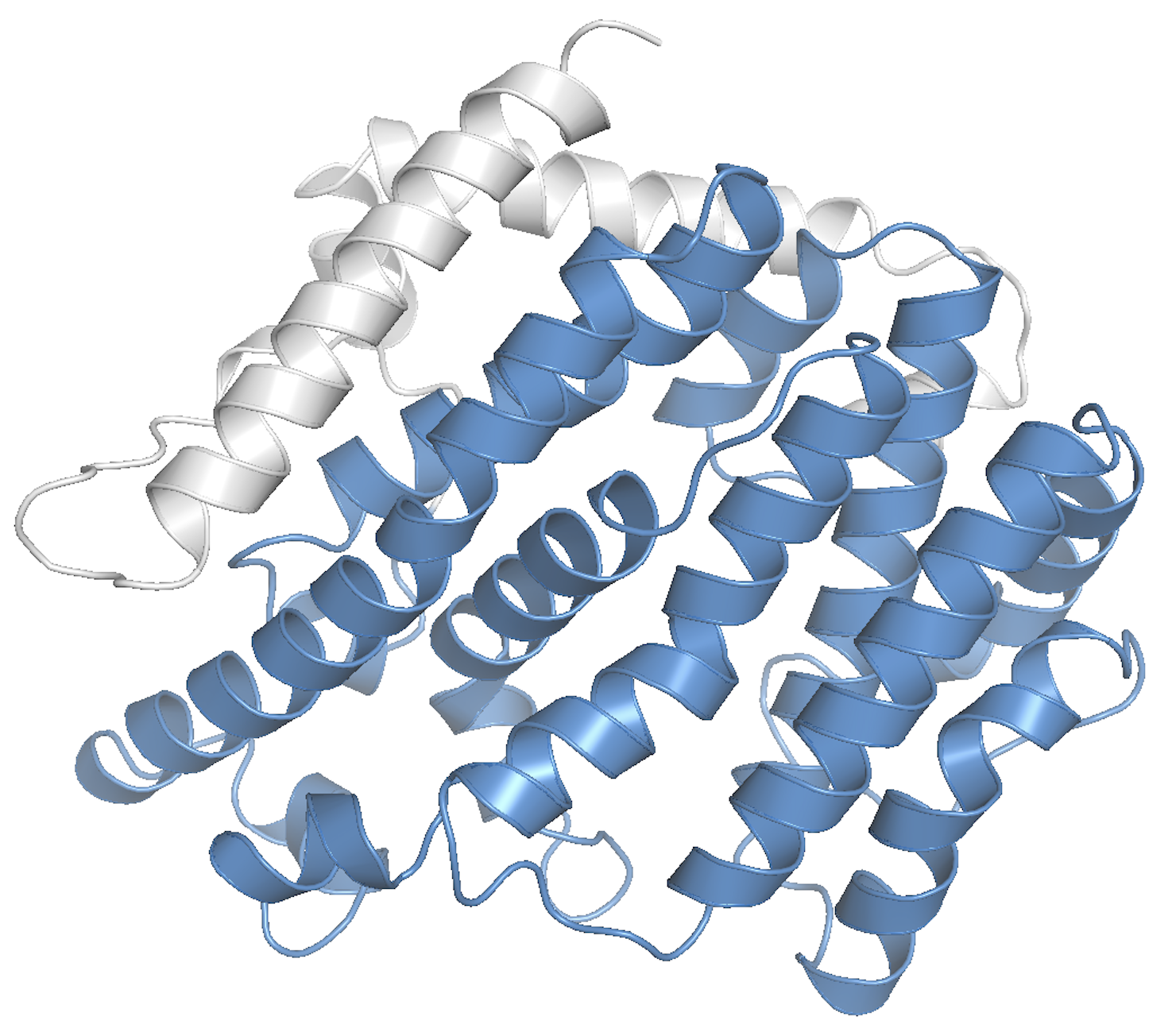}
        \caption*{TpsGPT2}
        \caption*{maxID 59.72\%}
    \end{subfigure}\hfill
    \begin{subfigure}[b]{0.16\textwidth}
        \includegraphics[width=\linewidth]{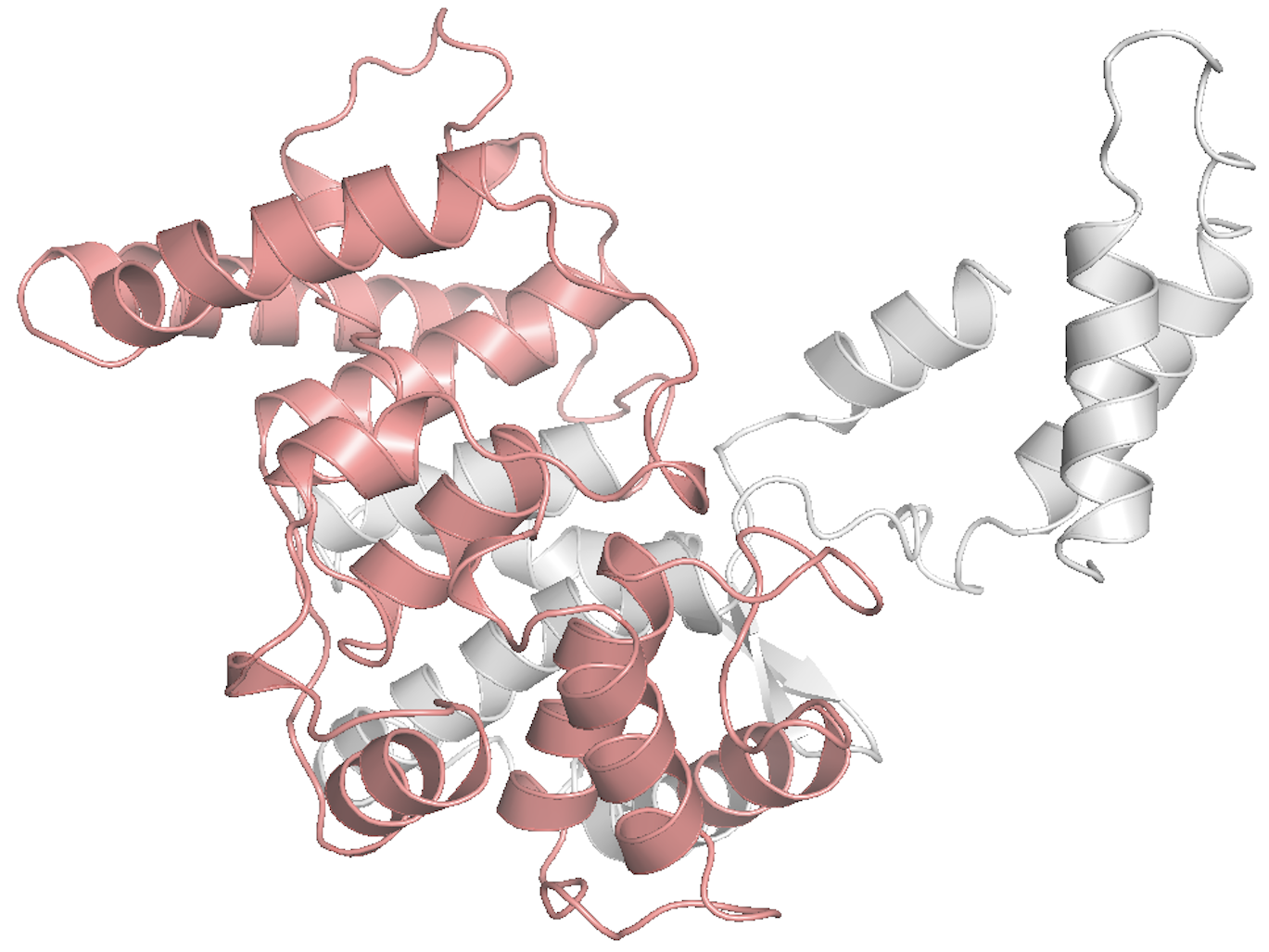}
        \caption*{TpsGPT3}
        \caption*{maxID 60.00\%}
    \end{subfigure}\hfill
    \begin{subfigure}[b]{0.16\textwidth}
        \includegraphics[width=\linewidth]{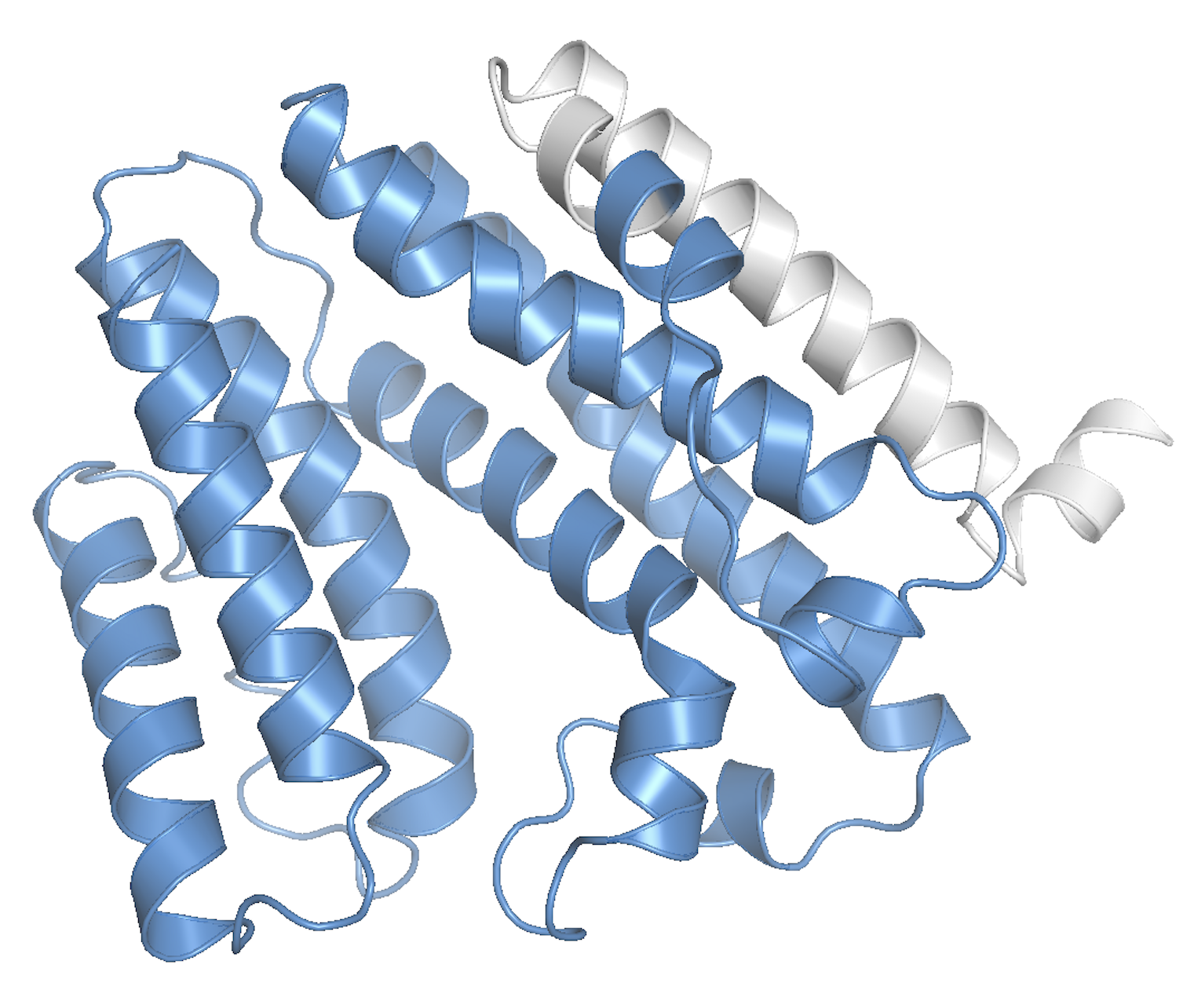}
        \caption*{TpsGPT4}
        \caption*{maxID 60.08\%}
    \end{subfigure}
\end{center}


  \begin{center}
    \begin{subfigure}[b]{0.16\textwidth}
      \includegraphics[width=\linewidth]{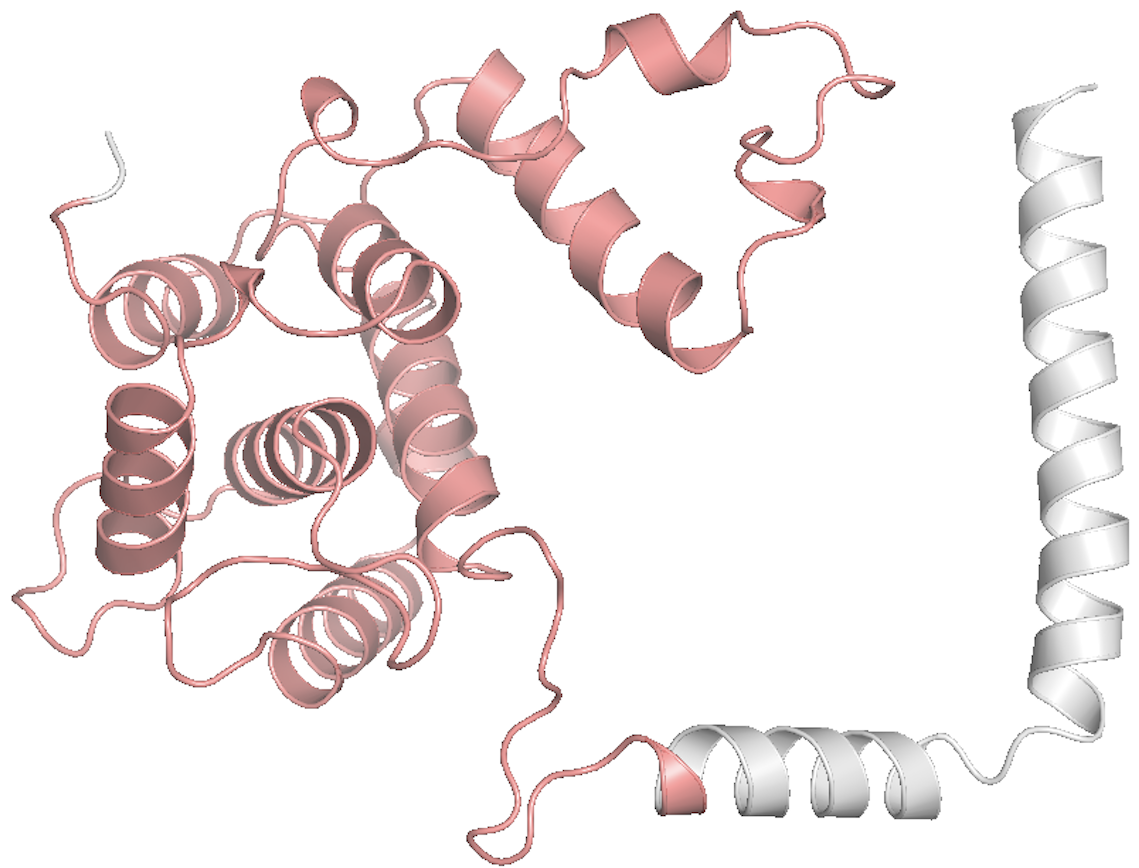}
      \caption*{TpsGPT5}
      \caption*{maxID 59.75\%}
    \end{subfigure}
    \hspace{0.06\textwidth}
    \begin{subfigure}[b]{0.16\textwidth}
      \includegraphics[width=\linewidth]{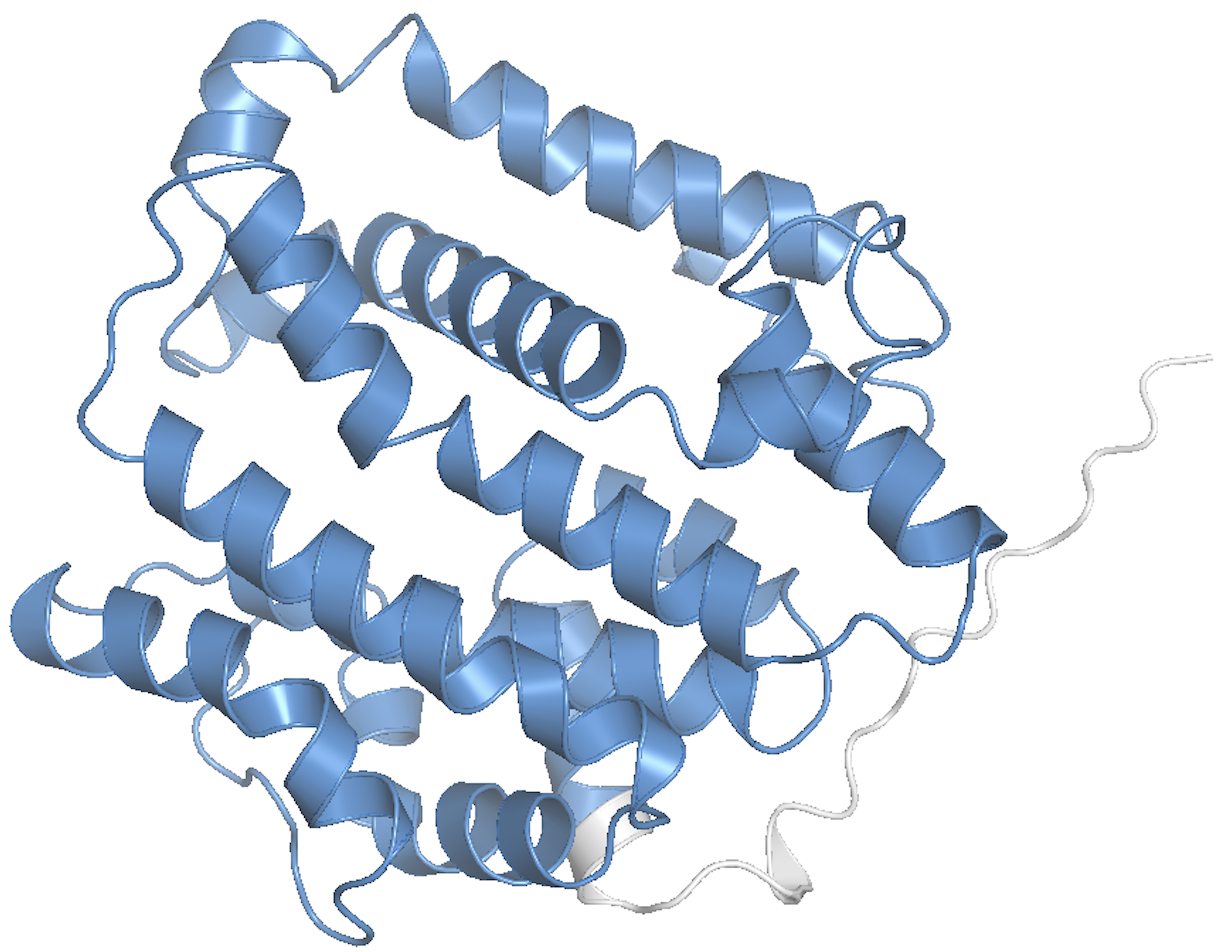}
      \caption*{TpsGPT6}
      \caption*{maxID 57.33\%}
    \end{subfigure}
    \hspace{0.06\textwidth}
    \begin{subfigure}[b]{0.16\textwidth}
      \includegraphics[width=\linewidth]{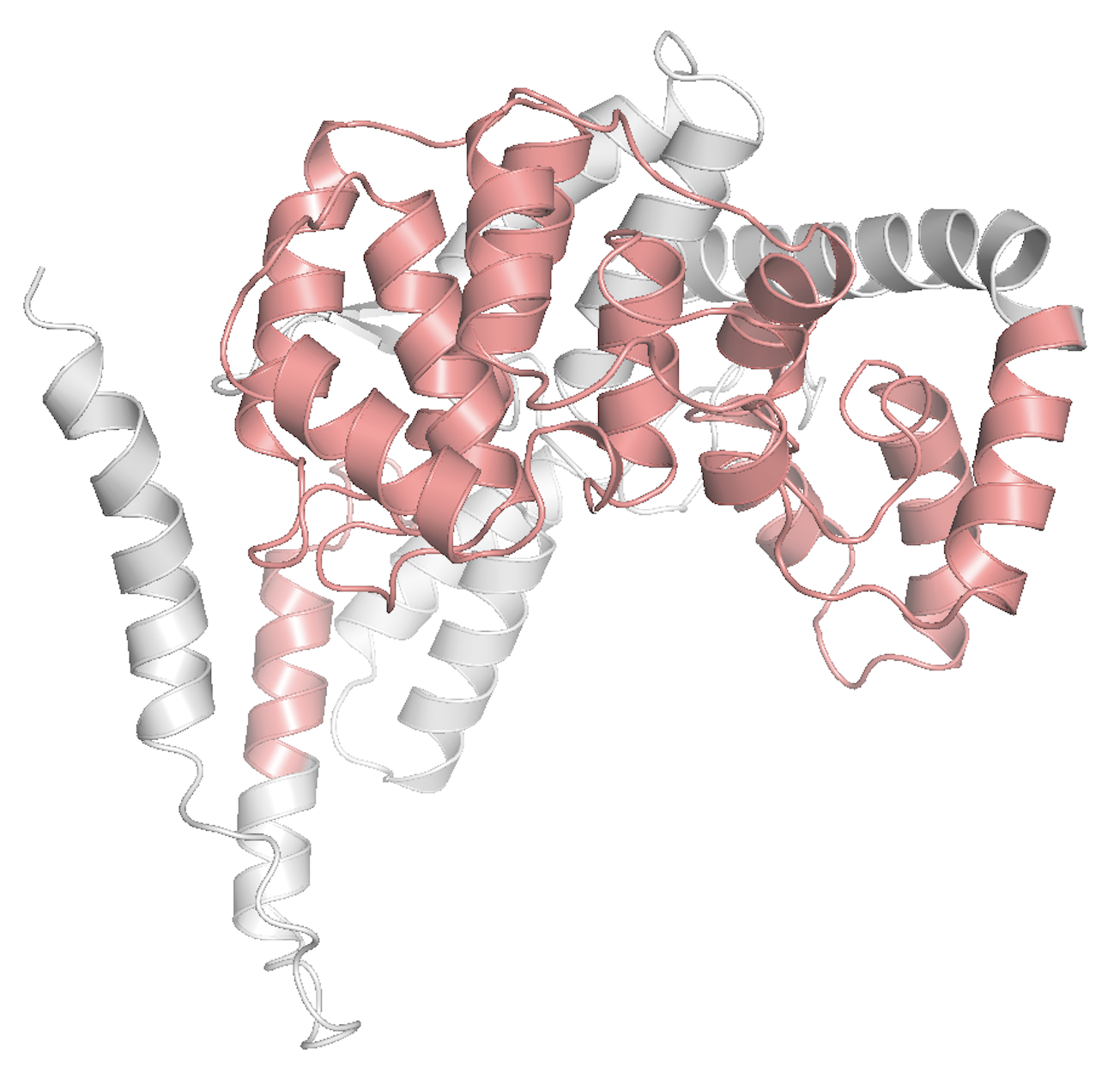}
      \caption*{TpsGPT7}
      \caption*{maxID 52.19\%}
    \end{subfigure}
    \begin{flushleft}
    \begin{subfigure}[t]{0.6\textwidth}
    \includegraphics[width=0.7\textwidth]{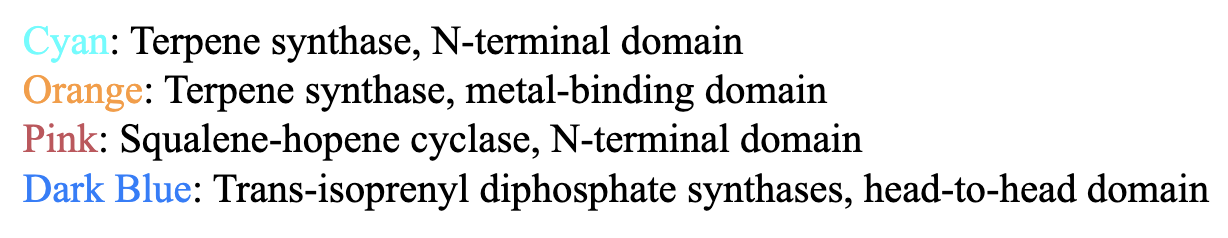}
    \end{subfigure}
    \end{flushleft}
  \end{center}
  
\caption{ColabFold-generated 3D structures of the seven \textit{de novo} putative TPS enzymes, with TPS domains annotated by InterPro~\citep{mirdita-2022, blum-2024}. maxID denotes the maximum sequence identity of each sequence to those in the training set. The figure shows that TpsGPT can generate evolutionarily distant TPS sequences while conserving TPS domains.}

\label{fig:all_tps}
\end{figure}

\vspace{-0.25cm}
From the 77 candidates, we filtered down to seven with $\leq 60\%$ sequence identity to the training set, representing potential \textit{de novo} TPS enzymes (Table~\ref{table:scores}). 3D structural comparison of the generated sequences relative to their respective top structural matches in the training set using Foldseek confirmed TM-scores between 0.6 and 0.9 (Table~\ref{table:scores}), consistent with belonging to the same structural family~\citep{van-kempen-2023}. Moreover, CLEAN assigned all seven sequences to TPS EC classes, providing robust computational support~\citep{yu-2023} (Table~\ref{table:scores}). InterPro analysis detected at least one relevant TPS specific domain in each sequence as shown in Figure~\ref{fig:all_tps}~\citep{blum-2024}. Together, the above results show that TpsGPT generates evolutionary distant yet structurally conserved \textit{de novo} TPS enzymes.

\vspace{-1em}

\smallskip

\begin{table}[H]
\centering
\caption{Properties of the seven putative \textit{de novo} TPS sequences. Each sequence is distinguished by a unique Sequence ID.  EnzymeExplorer TPS score measures TPS-like characteristics, pLDDT score indicates the stability of 3D folding, Max Foldseek TM-score indicates structural alignment with TPS sequences in training set, Max seq. identity to training set denotes the uniqueness of the TPS sequence, CLEAN-predicted EC number provides the enzyme classification, and InterPro predicts the domains in the sequence. TpsGPT1 and TpsGPT2 are shown in bold, indicating experimentally-validated enzymatic activity in both sequences.}

\label{table:scores}
\vspace{0.5em}

\resizebox{\textwidth}{!}{%
\begin{tabular}{c c c c c >{\RaggedRight\arraybackslash}p{3.5cm} >{\RaggedRight\arraybackslash}p{4.5cm}}
\toprule
\textbf{Sequence ID} &
\textbf{EnzymeExplorer TPS} &
\textbf{pLDDT} &
\textbf{Max Foldseek TM-Score} &
\textbf{Max seq. identity} &
\textbf{CLEAN Predicted} &
\textbf{InterPro Predicted} \\
 & 
\textbf{Score} &
\textbf{Score} &
\textbf{to training set} &
\textbf{to training set} & 
\textbf{EC number}&
\textbf{Domain}
 \\
\midrule
\textbf{TpsGPT1} & 0.75 & 78 & 0.73 & 49.67\% & Germacrene D Synthase (4.2.3.75) & Terpene synthase, N-terminal domain and Terpene synthase, metal-binding domain\\
\textbf{TpsGPT2} & 0.72 & 74 & 0.79 & 59.72\% & Squalene Synthase (2.5.1.21) & Trans-isoprenyl diphosphate synthases, head-to-head domain \\
TpsGPT3 & 0.73 & 74 & 0.84 & 60.00\% & Cucurbitadienol Synthase (5.4.99.33) & Squalene-hopene cyclase, N-terminal domain \\
TpsGPT4 & 0.73 & 70 & 0.65 & 60.08\% & Squalene Synthase (2.5.1.21) & Trans-isoprenyl diphosphate synthases, head-to-head domain \\
TpsGPT5 & 0.78 & 80 & 0.72 & 59.75\% & Beta-amyrin Synthase (5.4.99.39) & Squalene-hopene cyclase, N-terminal domain \\
TpsGPT6 & 0.73 & 71 & 0.69 & 57.33\% & Squalene Synthase (2.5.1.21) & Trans-isoprenyl diphosphate synthases, head-to-head domain \\
TpsGPT7 & 0.74 & 71 & 0.72 & 52.19\% & Cycloartenol Synthase (5.4.99.8)& Squalene-hopene cyclase, N-terminal domain \\
\bottomrule
\end{tabular}%
}
\end{table}

\vspace{-1.5em}
\subsection{Experimental Validation Confirms Enzymatic Activity}
\begin{figure}[H]
  \centering
  \includegraphics[width=0.58\textwidth]{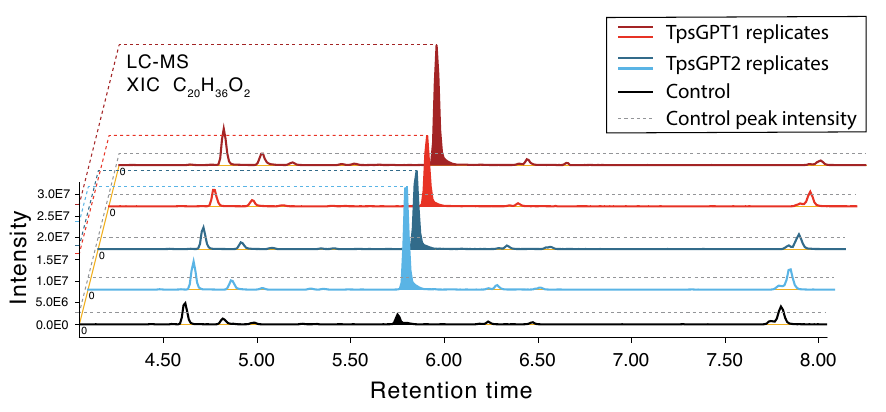}
  \caption{Chromatogram showing wet-lab validation of enzymatic activity for the generated \textbf{TpsGPT1} and \textbf{TpsGPT2} sequences. Extracted ion chromatograms (XIC) at the mass of C$_{20}$H$_{36}$O$_{2}$ confirm the production of diterpene-like products (e.g., sclareol; CHEBI:9053) in yeast expressing TpsGPT1 (red; two replicates) and TpsGPT2 (blue; two replicates). Black trace represents the control.}
  \label{fig:chromatogram}
  \vspace{-1.5em}
\end{figure}

\vspace{-0.75em}
To functionally characterize the enzymes designed with TpsGPT, we heterologously expressed the corresponding genes in the budding yeast Saccharomyces cerevisiae strain JWY501. This strain has been engineered for elevated production of the diterpene substrate geranylgeranyl pyrophosphate. Using liquid chromatography, coupled with mass spectrometry (LC-MS)~\citep{pitt-2009}, we confirmed the enzymatic activity in two sequences (\textbf{TpsGPT1} and \textbf{TpsGPT2}) (Figure~\ref{fig:chromatogram}).

\section{Conclusion}
\vspace{-0.75em}
In this work, we demonstrate the potential of fine-tuning protein language models, specifically ProtGPT2 Tiny, on a carefully curated TPS dataset to generate novel and valid terpene synthases. The seven sequences generated by \textbf{TpsGPT} exhibited high pLDDT scores, indicating stable 3D structures, and low perplexity scores, suggesting syntactically valid protein sequences. The seven sequences were also functionally validated by EnzymeExplorer, CLEAN and InterPro models. Furthermore, low pairwise sequence identity and favorable Foldseek TM-scores indicate the likelihood of discovering evolutionarily distant TPS enzymes not present in nature. Importantly, the entire pipeline was executed with less than \textbf{\$200} in GPU cost, demonstrating the scalable and cost-efficient nature of this approach. Additionally, our approach can also be applied to underrepresented protein families with little characterized enzyme datasets. These results validate our hypothesis that ProtGPT2 can be fine-tuned on a carefully curated TPS dataset to produce valid \textit{de novo} TPS candidates.


Among the seven generated sequences, enzymatic activity was so far confirmed in only two, and the presence of oxygen in the product chemical formula suggests they cannot yet be confirmed as canonical TPS enzymes. Ongoing experiments aim to characterize their catalytic mechanisms in more detail, as well as to validate other generated TPSs and it will help refine our \textit{in silico} pipeline. Future directions include conditioning TPS generation on terpene subclasses via curated datasets to generate specific terpenes and generalizing the methodology to other protein families, such as, for example, lysozymes, to explore functional diversity.

\section{Acknowledgments}

T.P. was supported by the European Union's Horizon Europe program (ERC, TerpenCode, 101170268 and Marie Skłodowska-Curie Actions, ModBioTerp, 101168583). J.S. was supported by the European Union's Horizon Europe program (ERC, FRONTIER, 101097822, ELIAS, 101120237 and CLARA, 101136607). Views and opinions expressed are however those of the author(s) only and do not necessarily reflect those of the European Union or the European Research Council. Neither the European Union nor the granting authority can be held responsible for them.

\bibliography{refs}

\section{Appendix}

\setcounter{table}{0}
\renewcommand{\thetable}{A\arabic{table}}
\setcounter{figure}{0}
\renewcommand{\thefigure}{A\arabic{figure}}

\subsection{Supplementary Methods}

\subsubsection{Hyperparameter Optimization}

\begin{figure}[H]
    \begin{minipage}{0.48\textwidth}
        \includegraphics[width=\linewidth]{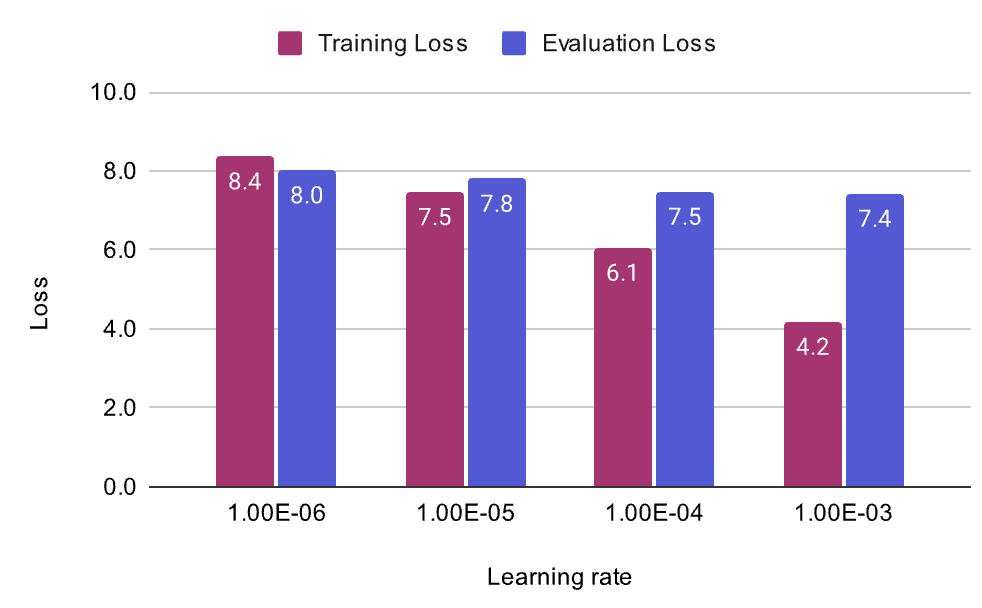}
        \caption{Training and evaluation loss as a function of learning rate.}
        \label{fig:lr_losses}
    \end{minipage}
    \hfill
    \begin{minipage}{0.48\textwidth}
        \captionof{table}{Training and evaluation loss for different learning rates.}
        \begin{tabular}{c c c}
            \toprule
            Learning rate & Training Loss & Evaluation Loss \\
            \midrule
            1e-6 & 8.4 & 8.0 \\
            1e-5 & 7.5 & 7.8 \\
            \textbf{1e-4} & \textbf{6.1} & \textbf{7.5} \\
            1e-3 & 4.2 & 7.4 \\
            \bottomrule
        \end{tabular}
        \label{tab:lr_losses}
    \end{minipage}
\end{figure}

\begin{figure}[H]
    \begin{minipage}{0.48\textwidth}
        \includegraphics[width=\linewidth]{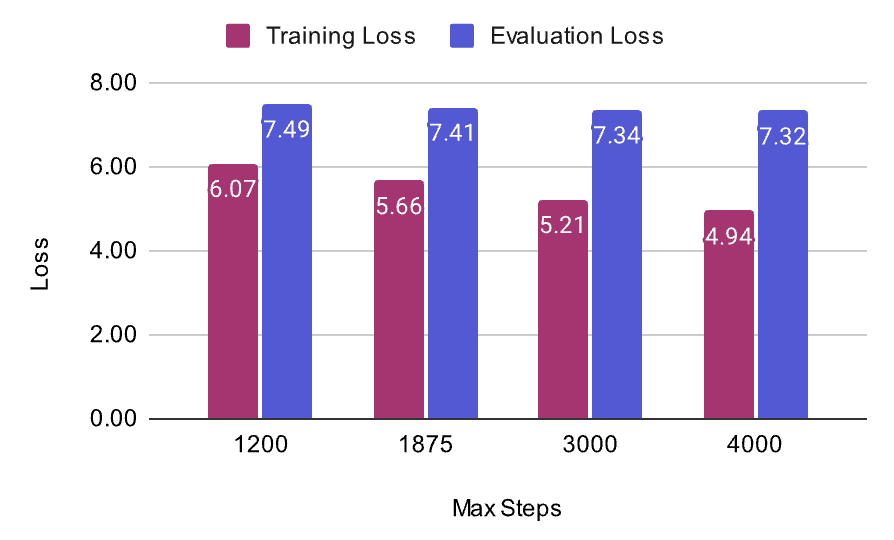}
        \caption{Training and evaluation loss as a function of max steps.}
        \label{fig:maxsteps_losses}
    \end{minipage}
    \hfill
    \begin{minipage}{0.48\textwidth}
        \captionof{table}{Training and evaluation loss as a function of max steps.}
        \begin{tabular}{c c c}
            \toprule
            Max steps & Training Loss & Evaluation Loss \\
            \midrule
            1200  & 6.07 & 7.49 \\
            1875  & 5.66 & 7.41 \\
            3000  & 5.21 & 7.34 \\
            \textbf{4000} & \textbf{4.94} & \textbf{7.32} \\
            \bottomrule
        \end{tabular}
        \label{tab:maxsteps_losses}
    \end{minipage}
\end{figure}

To obtain a well-generalized model, we optimized key hyperparameters of the ProtGPT2 fine-tuning process using the \textit{run\_clm.py} script from HuggingFace. The following hyperparameters were considered:

\begin{enumerate}

\item \textbf{Learning rate:} Learning rate controls how quickly the model updates its weights based on the training sequences. After experimentation, we selected a learning rate of 1e-4, at which point the validation loss converged. Higher rates (e.g., 1e-3) resulted in continued training loss reduction but increased overfitting, as shown in Table \ref{tab:lr_losses} and Figure \ref{fig:lr_losses}.

\item \textbf{Block size:} We used a block size of 512 tokens, consistent with the original ProtGPT2 paper. Each block represents the maximum sequence length fed into the model during training.

\item \textbf{Batch size and Gradient accumulation steps:} Regular batch size was set to 64. To simulate a larger effective batch size of 512 on a single GPU, we set \textbf{gradient accumulation steps} to 8, summing the gradients over eight steps during backpropagation.

\item \textbf{Max steps:} The max steps parameter controls the total number of optimization steps (analogous to training epochs). We empirically determined that 4,000 steps were optimal, achieving convergence in both training and validation loss (Table \ref{tab:maxsteps_losses} and Figure \ref{fig:maxsteps_losses}).

\end{enumerate}

\subsubsection{Sequence Validation Methods}

\begin{figure}[H]
  \centering
  \begin{center}
    \begin{subfigure}[b]{0.18\textwidth}
        \includegraphics[width=\linewidth]{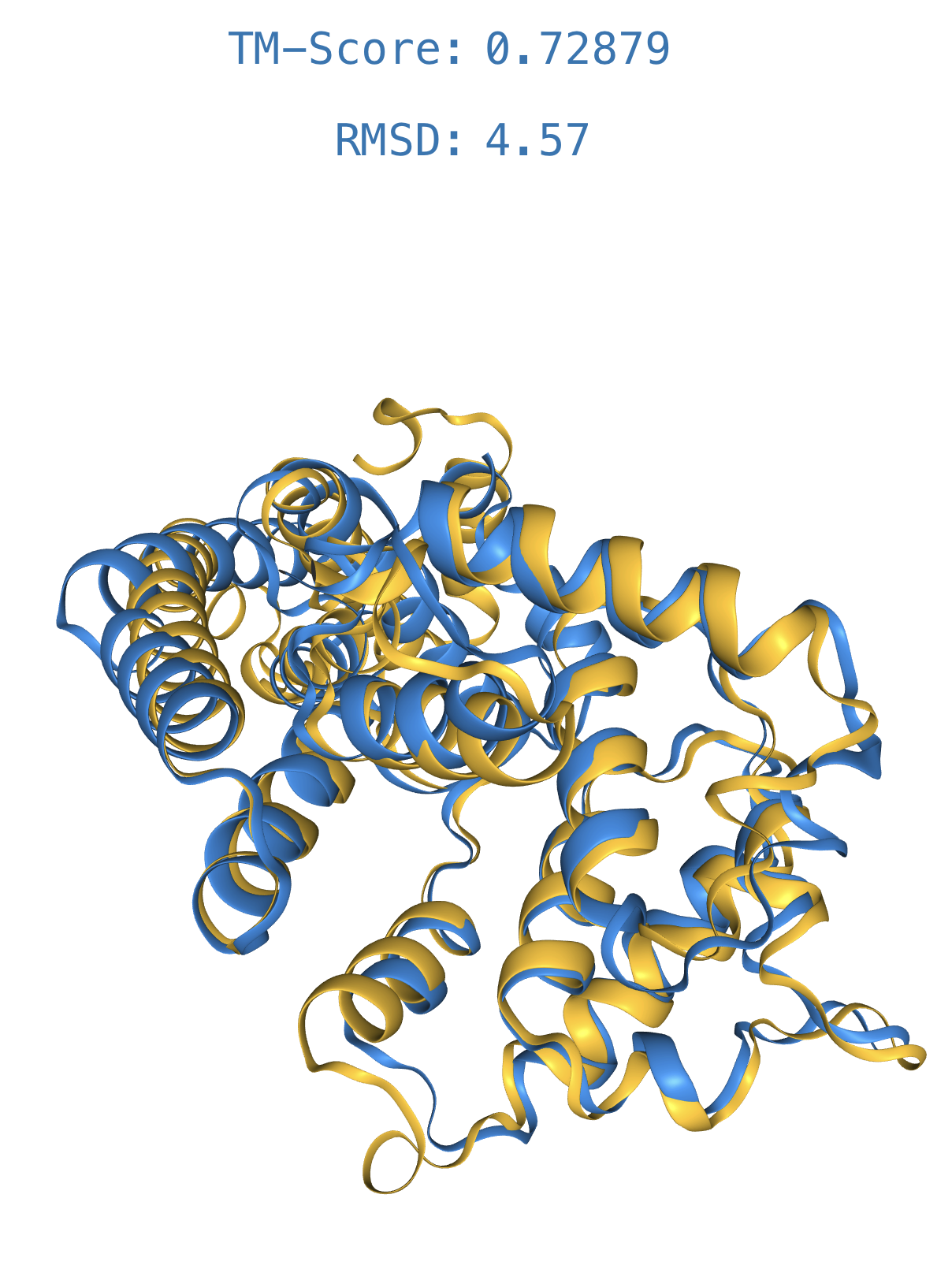}
        \caption*{TpsGPT1}
    \end{subfigure}\hfill
    \begin{subfigure}[b]{0.18\textwidth}
        \includegraphics[width=\linewidth]{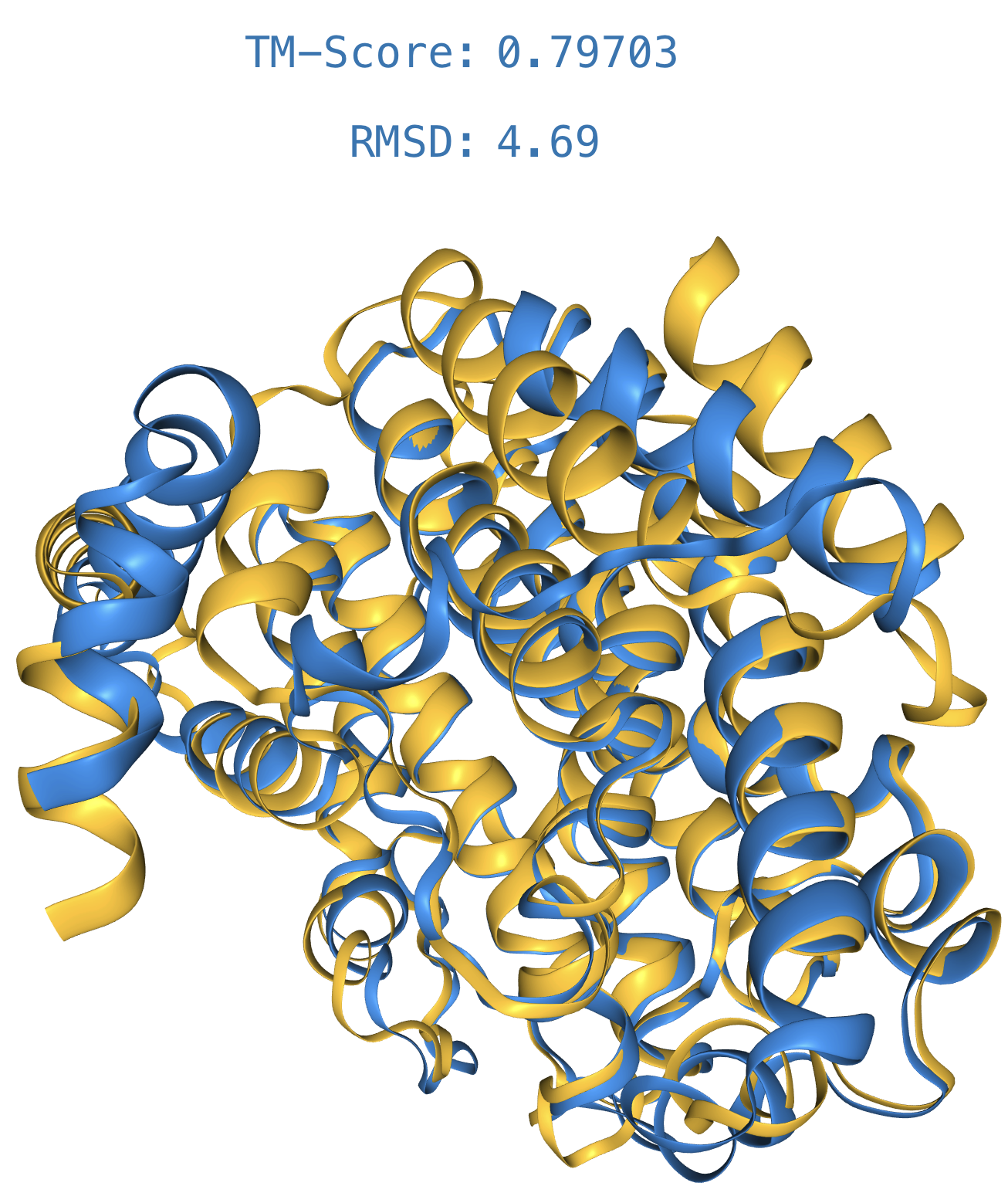}
        \caption*{TpsGPT2}
    \end{subfigure}\hfill
    \begin{subfigure}[b]{0.18\textwidth}
        \includegraphics[width=\linewidth]{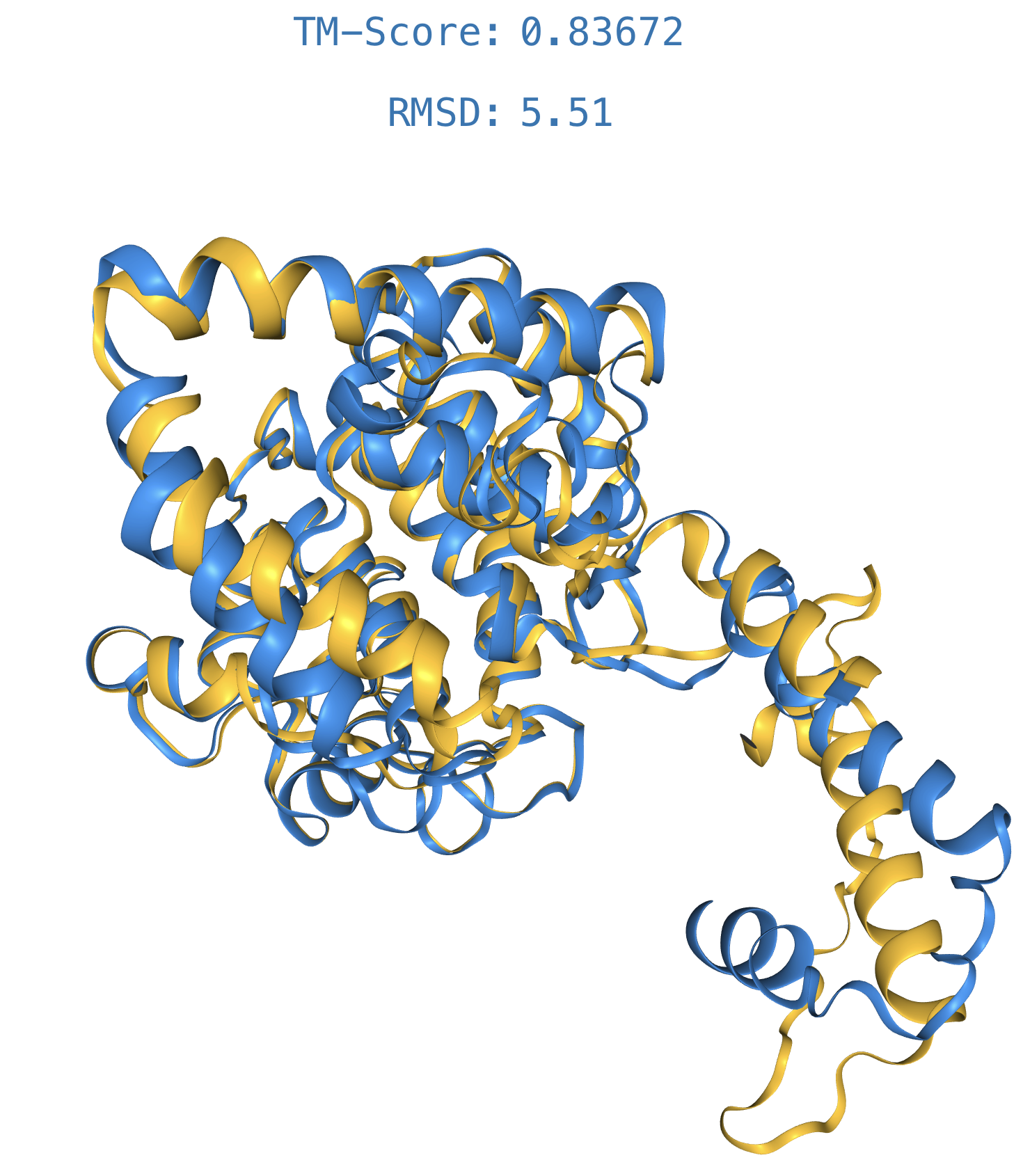}
        \caption*{TpsGPT3}
    \end{subfigure}\hfill
    \begin{subfigure}[b]{0.18\textwidth}
        \includegraphics[width=\linewidth]{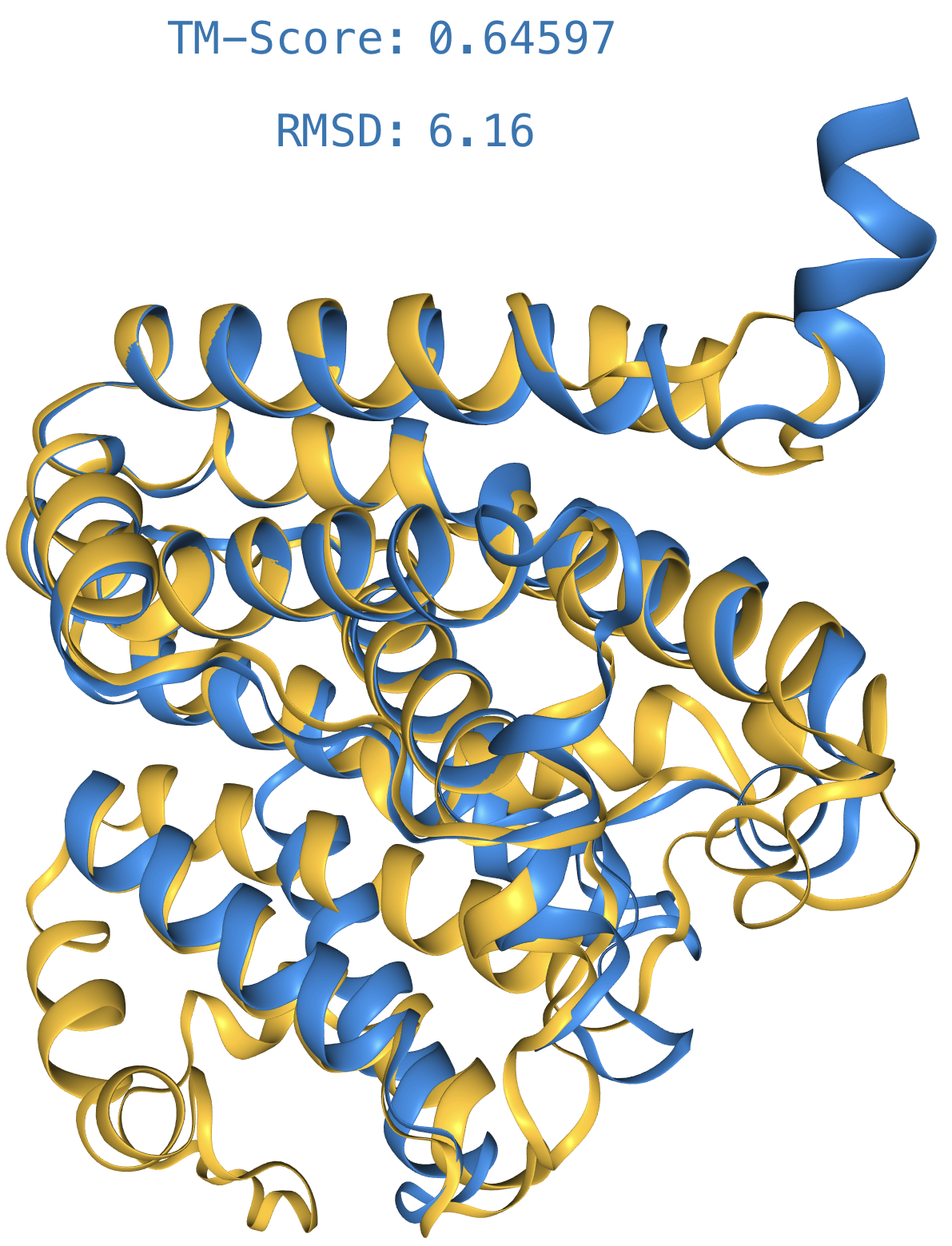}
        \caption*{TpsGPT4}
    \end{subfigure}
\end{center}

  \begin{center}
    \begin{subfigure}[b]{0.18\textwidth}
      \includegraphics[width=\linewidth]{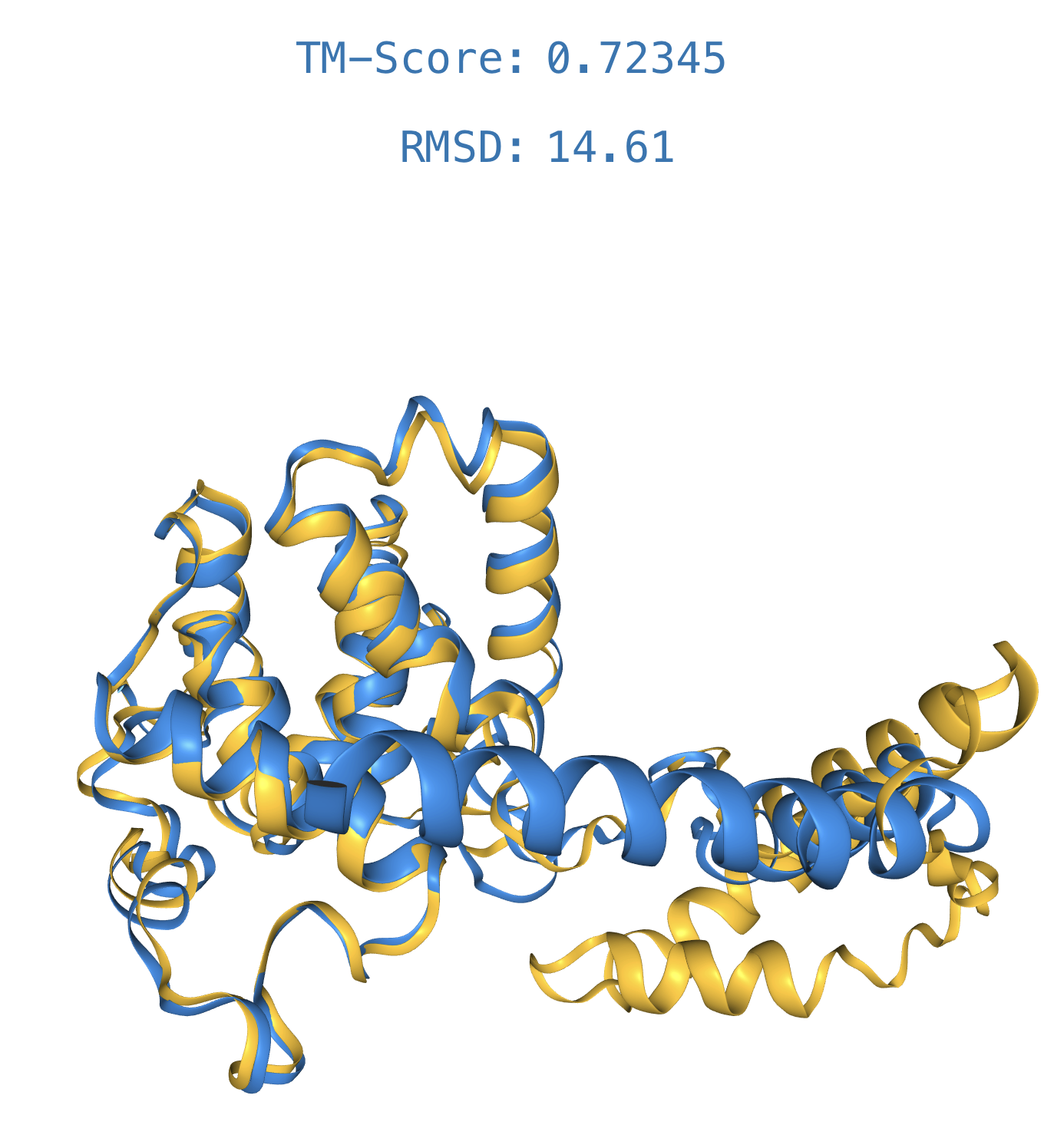}
      \caption*{TpsGPT5}
    \end{subfigure}
    \hspace{0.06\textwidth}
    \begin{subfigure}[b]{0.18\textwidth}
      \includegraphics[width=\linewidth]{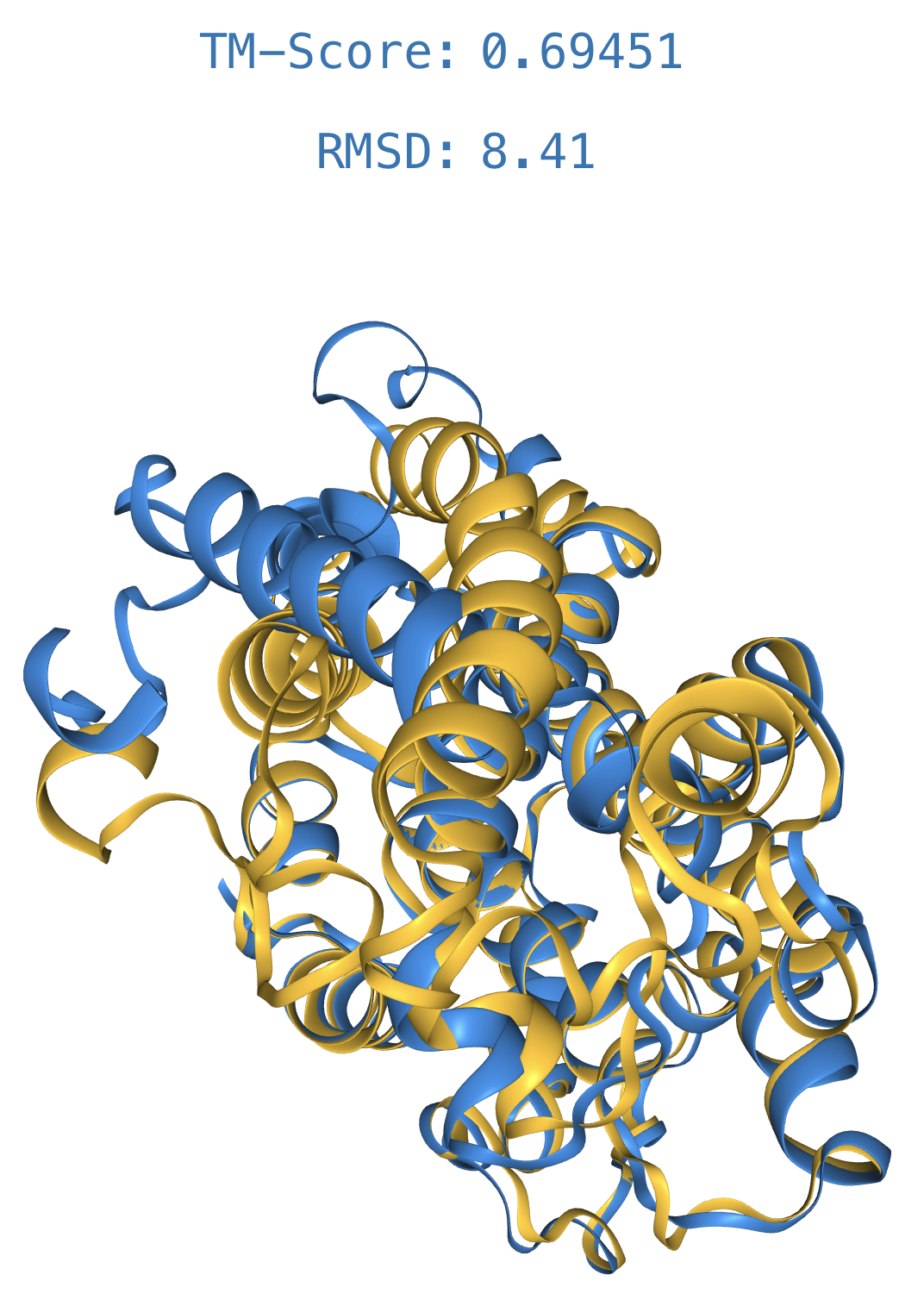}
      \caption*{TpsGPT6}
    \end{subfigure}
    \hspace{0.06\textwidth}
    \begin{subfigure}[b]{0.18\textwidth}
      \includegraphics[width=\linewidth]{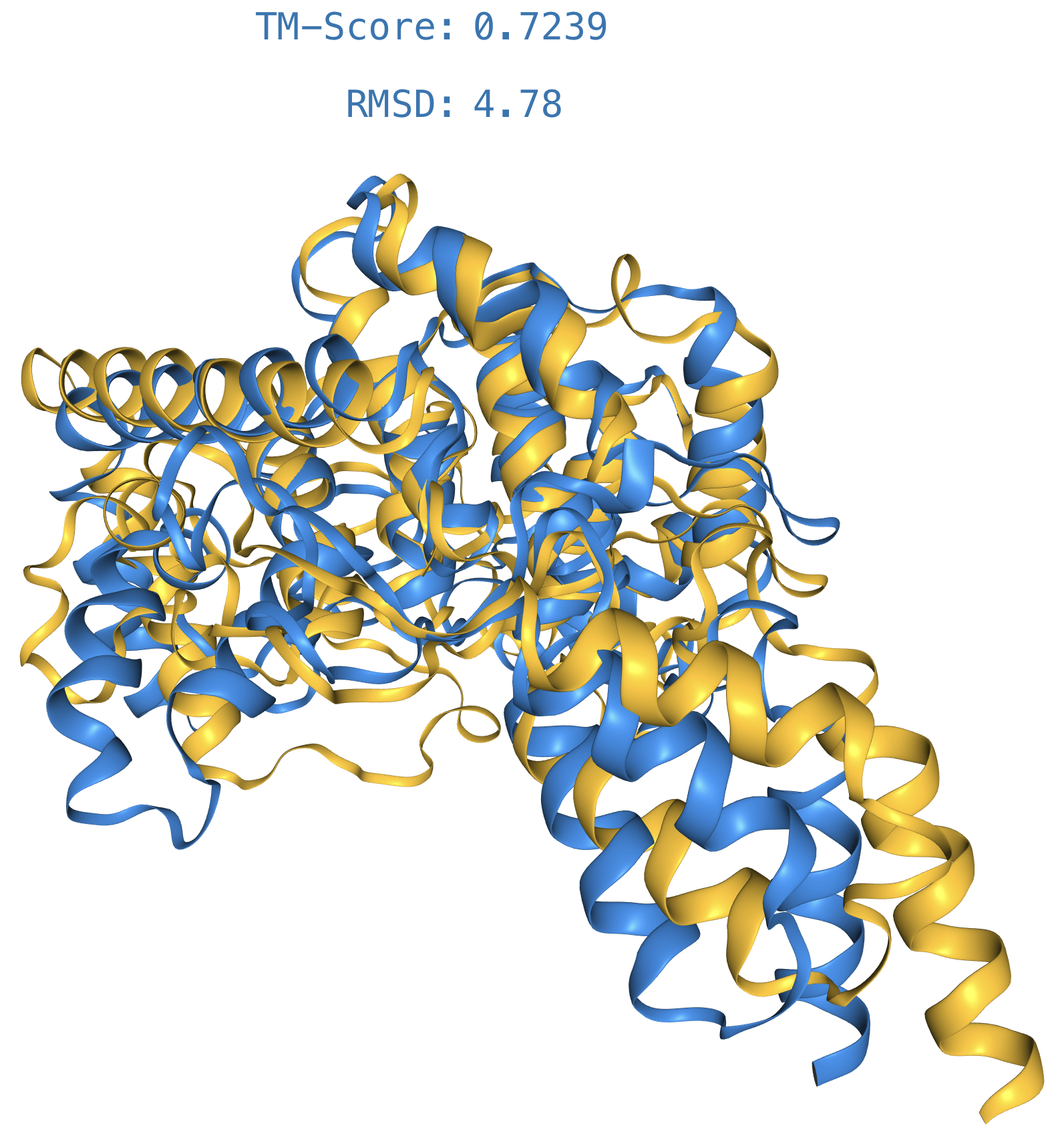}
      \caption*{TpsGPT7}
    \end{subfigure}
  \end{center}
  
\caption{Foldseek structural alignment for the seven TPS sequences with their respective top matches in the training set. Foldseek TM-scores were between 0.6 and 0.9 consistent with belonging to the same TPS family. Blue represents the generated TPS sequences and yellow is the target top structural match in the training set.}
\label{fig:foldseek_alignment}
\end{figure}

\begin{enumerate}

\item \textbf{EnzymeExplorer}: We applied the EnzymeExplorer command-line tools~\citep{pluskal-lab-enzyme-explorer} with a detection threshold of 0.7 to identify putative TPS sequences.

\item \textbf{CLEAN}: We used the CLEAN web server~\citep{clean-server} to predict the EC numbers for the seven generated TPS sequences.

\item \textbf{InterPro}: We used the InterProScan web server~\citep{interpro-server} to predict the protein domains for the seven generated TPS sequences.

\item \textbf{Foldseek}: We employed the Foldseek command-line tools~\citep{steineggerlab} to construct a target database from the training set proteins (63k). Using the \textit{easy-search} command, we identified the top structural match in this database for each of the seven generated TPS sequences and recorded the corresponding TM-score (Figure \ref{fig:foldseek_alignment}).

\end{enumerate}

\subsection{Threshold Selection}

\begin{enumerate}

\item \textbf{maxID}: maxID threshold of $\leq$ 60\% was chosen based on past experimental data from protein generation~\citep{ruffolo-2025}.

\item \textbf{pLDDT}: We chose a pLDDT threshold of $\geq$ 70 which corresponds to a correct backbone prediction with misplacement of some side chains~\citep{embl-ebi-no-date}

\item \textbf{EnzymeExplorer Threshold}: EnzymeExplorer TPS detection threshold was $\geq$ 0.7 to identify likely TPS sequences. Past research work has used thresholds between 0.35 and 0.7~\citep{samusevich-2024}.

\item \textbf{TM-Score}: We filtered to sequences with TM-Scores between 0.6 and 0.9 to preserve structural similarity and fold with representative TPS functions.

\end{enumerate}

\subsection{Appendix Tables and Figures}

\vspace{-1cm}
\begin{figure}[H]
  \centering
    \includegraphics[width=0.5\linewidth]{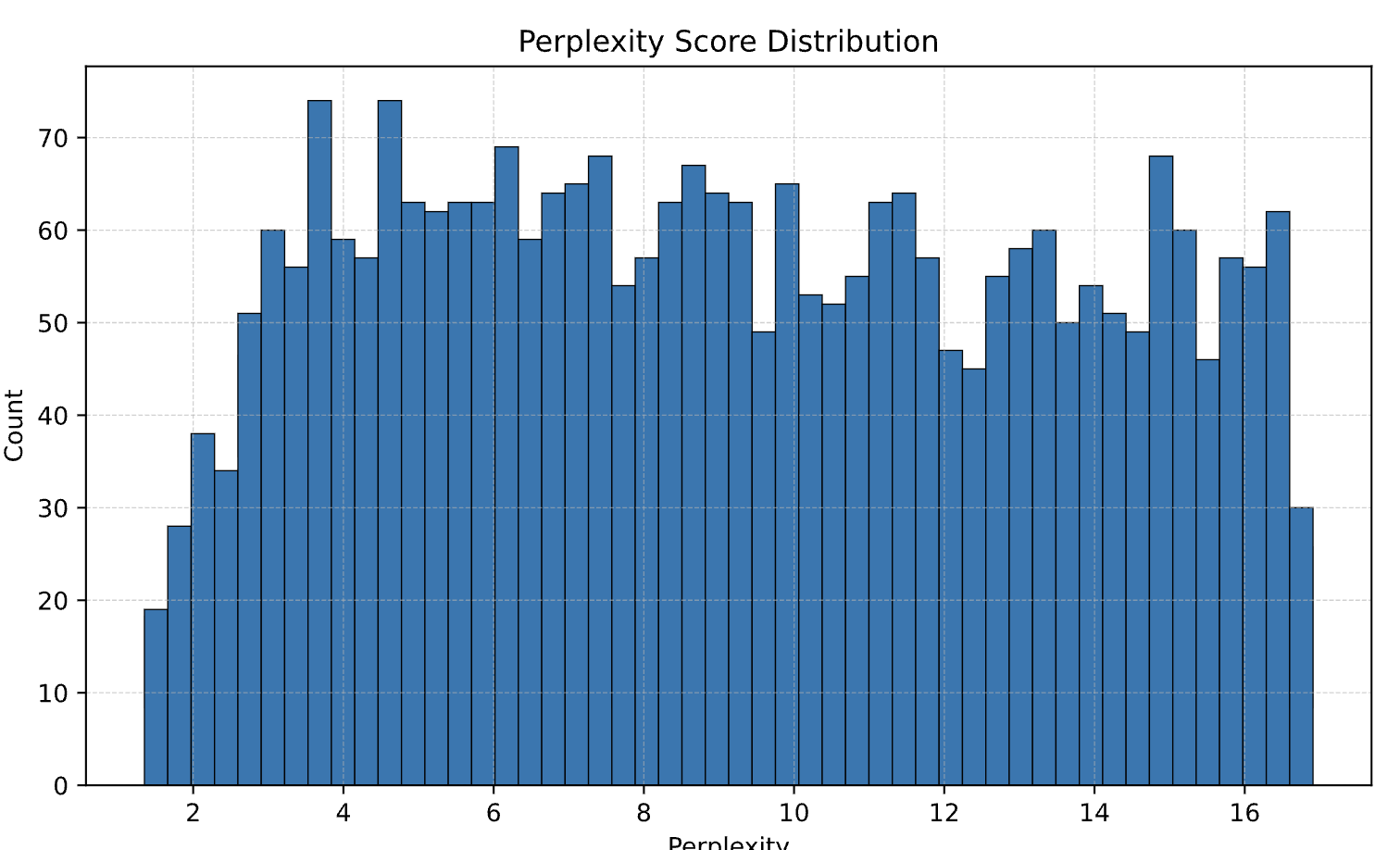}
  \caption{Distribution of perplexity scores for the top 2800 sequences.}
  \label{fig:perplexity_scores}
\end{figure}

\end{document}